\documentclass[runningheads]{llncs}
\usepackage[T1]{fontenc}
\usepackage{graphicx}
\usepackage{marvosym}
\usepackage[ruled, noend, noline]{algorithm2e}
\usepackage{algorithmic}
\usepackage{multirow}
\usepackage{threeparttable}
\usepackage{wrapfig}

\begin{document}

\title{MLF-DET: Multi-Level Fusion for Cross-Modal 3D Object Detection}
%
%
\author{
Zewei Lin
\and Yanqing Shen
\and Sanping Zhou\textsuperscript{(\Letter)}
\and Shitao Chen
\and Nanning Zheng
}
\authorrunning{Z. Lin et al.}

\institute{National Key Laboratory of Human-Machine Hybrid Augmented Intelligence, National Engineering Research Center for Visual Information and Applications, and Institute of Artificial Intelligence and Robotics, Xi’an Jiaotong University,
Xi’an, China\\
\email{
\{xianjiaoda2017zw,qing1159364090\}@stu.xjtu.edu.cn, 
\{spzhou, chenshitao,nnzheng\}@mail.xjtu.edu.cn}
}

\maketitle

\begin{abstract}
In this paper, we propose a novel and effective Multi-Level Fusion network, named as MLF-DET, for high-performance cross-modal 3D object DETection, which integrates both the feature-level fusion and decision-level fusion to fully utilize the information in the image. For the feature-level fusion, we present the Multi-scale Voxel Image fusion~(MVI) module, which densely aligns multi-scale voxel features with image features. For the decision-level fusion, we propose the lightweight Feature-cued Confidence Rectification~(FCR) module which further exploits image semantics to rectify the confidence of detection candidates. Besides, we design an effective data augmentation strategy termed Occlusion-aware GT Sampling~(OGS) to reserve more sampled objects in the training scenes, so as to reduce overfitting. Extensive experiments on the KITTI dataset demonstrate the effectiveness of our method. Notably, on the extremely competitive KITTI car 3D object detection benchmark, our method reaches 82.89\% moderate AP and achieves state-of-the-art performance without bells and whistles.

\end{abstract}

\section{Introduction}

3D object detection aims to get the object's location, size, and direction in 3D space, which are indispensable information for path planning and motion control in autonomous driving. The most frequently used sensors in 3D detection are LiDAR and camera. LiDAR point clouds can capture the structural and depth information, with low resolution and lack of semantic information. Instead, camera images provide rich semantic information, such as color and texture, while suffering from an inherent 3D information ambiguity. Considering the complementarity of point clouds and images, many works attempt to fuse the two sensor modalities to improve detection accuracy~\cite{vora2020pointpainting,huang2020epnet,zhu2022vpfnet,pang2020clocs,wu2022sparse,zhang2023bidirectional}. Nevertheless, how to overcome the modality gap and make good use of the two sensor data remains a tremendous challenge.

Currently, cross-modal 3D detection methods can be mainly divided into two lines, \emph{i.e.}, feature-level fusion ones and decision-level fusion ones. Feature-level fusion methods~\cite{vora2020pointpainting,huang2020epnet,zhu2022vpfnet,bai2022transfusion,zhang2022cat} use different backbone networks to extract the features of different modalities and merge the cross-modal features into a unified representation, which would be used for detection. Decision-level fusion methods~\cite{pang2020clocs,pang2022fast} use different detectors to process different modalities and ensemble the output boxes of the detectors to get the final results. Generally, the feature-level fusion methods can leverage the rich semantic features of images, but the complex fusion network is prone to overfitting and the image features may not be used properly. The decision-level fusion methods can use accurate 2D detection boxes to boost 3D detection performance but are short of fine-grained information interaction between two modalities.

In this paper, we present a novel 3D detection framework termed MLF-DET, which incorporates the advantages of feature-level fusion and decision-level fusion to fully exploit image information and achieve better performance. Specifically, instead of projecting LiDAR points onto the image plane and building sparse connections between point features and image features~\cite{vora2020pointpainting,huang2020epnet,liu2022epnet++}, the designed Multi-scale Voxel Image fusion~(MVI) module projects multi-scale voxel centers at non-empty locations onto the image plane. In this way, a dense connection between voxel features and image features would be built. That's because the regular sparse convolution network~\cite{graham20183d} dilates the sparse voxel features and produces new occupied voxels.
To further squeeze the juice of the image, the Feature-cued Confidence Rectification~(FCR) module treats the projected 3D RoIs as 2D RoIs and refines them to produce confidence scores, respectively. Afterwards, the confidence scores of 3D RoIs and 2D RoIs along with their features are fed into a lightweight MLP to produce the rectified scores. It is worth noting that this module does not introduce much complexity to the network due to the shared RoIs between the two modalities.

In addition, we also investigate cross-modal data augmentation to further improve detection performance. Data augmentation is essential for cross-modal networks for they contain more parameters and suffer from overfitting. The previous work~\cite{wang2021pointaugmenting} extends the commonly used GT Sampling data augmentation~\cite{yan2018second} from LiDAR-only methods to cross-modal methods. However, the severe occlusion on the image plane is ignored. Since the space on the image plane is quite limited, most sampled objects would be removed for they are heavily occluded with other objects, making GT Sampling less effective. Therefore, we propose Occlusion-aware GT Sampling~(OGS) to carefully deal with the occlusion of objects. By removing the most severely occluded sampled objects in order, we reserve more sampled image patches and alleviate overfitting in training.

Our main contributions can be summarized in three-fold.
(1)~We propose a novel multi-level fusion framework to fuse point clouds and images in the feature level and decision level to make full use of semantic information of images.
(2)~We present a novel occlusion-aware GT Sampling strategy to augment the data which can preserve more sampled objects and diversify the training scenes.
(3)~We experimentally validate the superiority of our method on the KITTI dataset~\cite{geiger2012we}. Without exploiting test time augmentation or model ensemble or extra data, we surpass state-of-the-art cross-modal methods with remarkable margins.

\section{Related Work}
\textbf{Camera-Only 3D Detection.}
Camera-only methods are rewarding because the cost of a camera is much lower than a LiDAR sensor. RTM3D~\cite{li2020rtm3d} predicts the nine perspective key points of a 3D bounding box in image space, and then utilizes the geometric constraints of perspective projection to estimate a stable 3D box. SMOKE~\cite{liu2020smoke} regresses 3D boxes directly with a single keypoint estimation and proposes a multi-step disentangling strategy to improve accuracy. MonoFlex~\cite{zhang2021objects} optimizes the detection performance of truncated objects by decoupling the edge of the feature map and predicts the depth by uncertainty-guided ensemble. Because of the inherent depth ambiguity of images, these methods suffer from low precision.

\textbf{LiDAR-Only 3D Detection.}
According to point cloud representations, LiDAR-only 3D detectors can be classified into  point-based ones and grid-based ones. The point-based detectors extract representative features directly from raw points with PointNet~\cite{qi2017pointnet} or its variant~\cite{qi2017pointnet++}. PointRCNN~\cite{shi2019pointrcnn} produces RoIs from foreground points in a bottom-up manner, which is followed by a refinement network. VoteNet~\cite{qi2019deep} learns to predict the instance centroids by introducing a novel deep Hough voting. IA-SSD~\cite{zhang2022not} argues that points are not equally important and presents two sampling strategies to preserve more foreground points. Grid-based detectors divide unstructured points into regular grids and extract features with Convolutional Neural Network~(CNN). SECOND~\cite{yan2018second} makes use of sparse convolutions to speed up the feature extraction of voxels. PointPillars~\cite{lang2019pointpillars} converts point clouds organized in pillars into pseudo images, so as to cut off the inefficient 3D convolutions. Voxel R-CNN~\cite{deng2021voxel} argues the precise positions of points are not needed and devises a voxel RoI pooling to extract RoI features. Since point clouds are very sparse, especially in the long range, the performance of LiDAR-only detectors is limited.

\textbf{Cross-Modal 3D Detection.}
To take advantage of both LiDAR and camera, many cross-modal methods are proposed. PointPainting~\cite{vora2020pointpainting} decorates each raw point with class scores from an image-based semantic segmentation network. EPNet~\cite{huang2020epnet} builds a LI-Fusion module to fuse point features with image features adaptively. EPNet++~\cite{liu2022epnet++} improves EPNet by introducing the CB-Fusion module and MC loss. A drawback of these approaches is the sparse correspondences between LiDAR points and image pixels, which leads to severe information loss of images. DVF~\cite{mahmoud2023dense} builds a dense voxel fusion pipeline by weighting each voxel feature with the foreground mask constructed from a 2D detector. However, the foreground mask of the image is  highly abstract and still causes information loss. SFD~\cite{wu2022sparse} generates pseudo point clouds from images with depth completion and fuses them with original point clouds in 3D space. CLOCs~\cite{pang2020clocs} fuses the candidates of a 3D detector and a 2D detector before NMS in the decision level. Fast-CLOCs~\cite{pang2022fast} reduces the memory usage and computational complexity of CLOCs by the 3D-Q-2D image detector, which shares 3D candidates with the image branch.  Nevertheless, the process of constructing the sparse tensor is still cumbersome and time-consuming.

\section{MLF-DET}
\begin{figure}[t!]
\includegraphics[width=\textwidth]{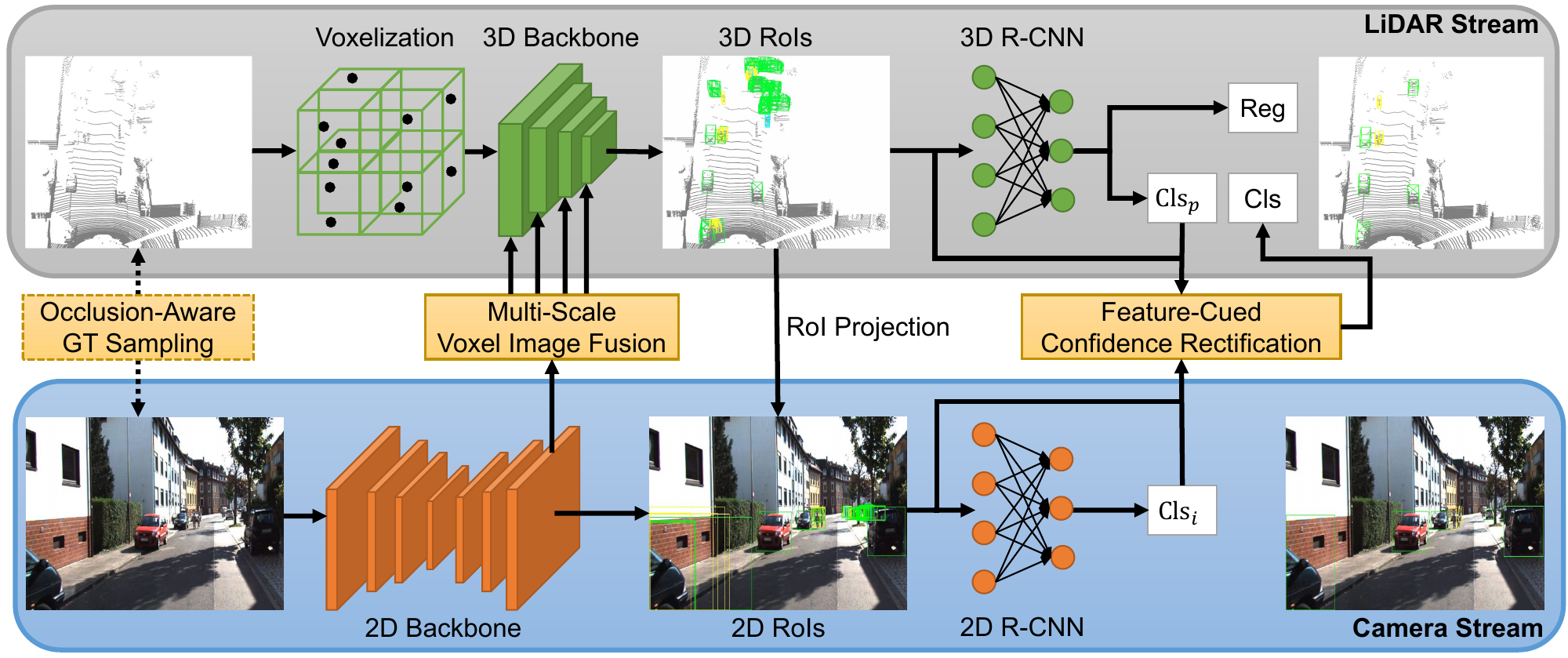}
\caption{Overview of MLF-DET. First, the Camera stream extracts the image features with ResNet and FPN. Second, the LiDAR stream voxelizes the point cloud and extracts voxel features with sparse convolutions. Simultaneously, the MVI module enhances the voxel features with image features. Then, the 3D RoIs generated from 3D RPN are shared with camera stream via projecting. Subsequently, the 3D and 2D RoIs are refined, respectively. Finally, the FCR module is applied to rectify the confidence of detection candidates. The OGS module shown in dashed lines is used to preserve more sampled objects in training and would be turned off in inference.} 
\label{architecture}
\end{figure}

In this section, we show the detailed design of MLF-DET. As illustrated in Fig.~\ref{architecture}, MLF-DET is composed of a LiDAR stream, a camera stream, and three modality interaction modules. MVI aims to utilize image semantic features to enhance voxel features and FCR plays a role in rectifying the confidence of detection candidates. As for OGS, it is used to reserve more sampled objects in training scenes. 

\subsection{Multi-Scale Voxel Image Fusion}
\begin{figure}[t!]
\includegraphics[width=\textwidth]{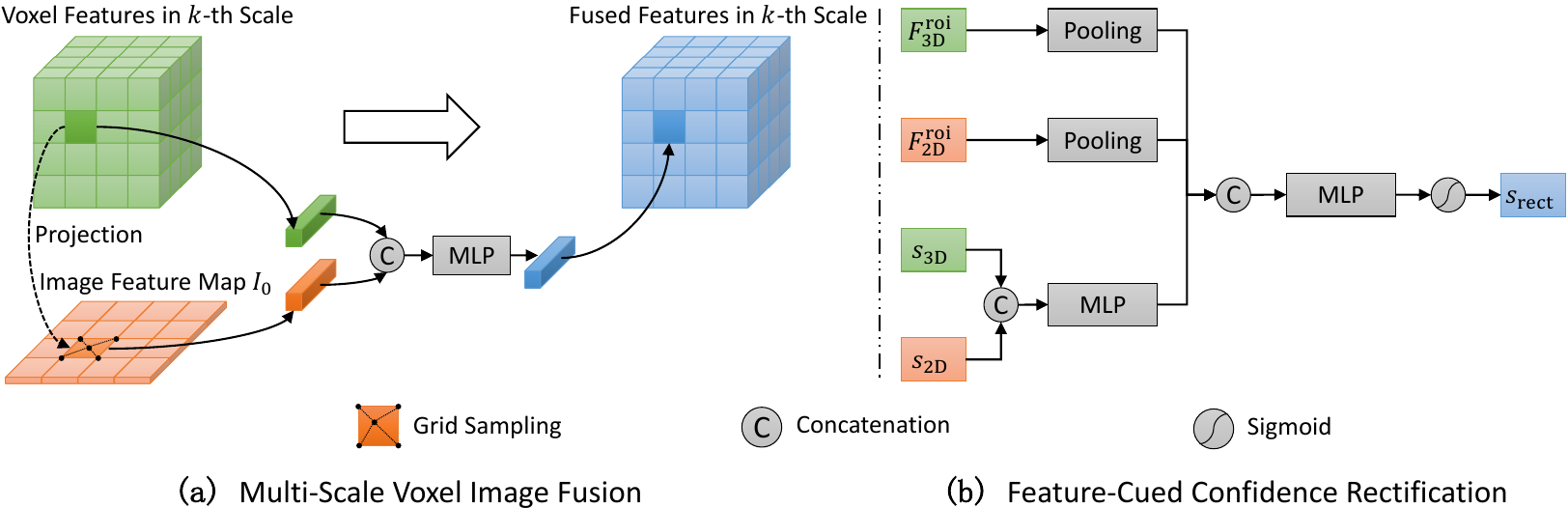}
\caption{Illustrations of (a) multi-scale voxel image fusion module which densely fuses multi-scale voxel features with image features, and (b) feature-cued confidence rectification module which rectify the confidences of RoIs using the pooled features of RoIs as cues.}
\label{feature_and_decision_fusion}
\end{figure}

In order to build a dense connection between point cloud and image and fully utilize image features to enrich voxel features, we devise the MVI module, as depicted in Fig~\ref{feature_and_decision_fusion}~(a). Specifically, we use ResNet50~\cite{he2016deep} to extract image features and further use FPN~\cite{lin2017feature} to build the feature pyramid \{$I_0, I_1, I_2, I_3, I_4$\}, where $I_0$ is the feature map with the finest level. Meanwhile, several 3D sparse convolution blocks are selected to extract multi-scale voxel features for voxelized point clouds. The features of occupied voxels in the $k$-th scale~($k \in \{0,1,2,3\}$) can be denoted by $V^k \in \mathbf{R}^{N_k \times C_k}$, where $N_k$ is the number of occupied voxels, $C_k$ is the number of feature channels. The centers of these voxels are denoted by $P^k \in \mathbf{R}^{N_k \times 3}$. For a particular voxel with features $v^k$ and center $p^k$, the corresponding position $p^k_{\rm{img}}$ on the image plane can be calculated as follows
\begin{equation}
    p^k_{\rm{img}} = T_{\rm{img} \gets \rm{cam}}T_{\rm{cam} \gets \rm{lidar}}p^k,
\end{equation}
where $T_{\rm{cam} \gets \rm{lidar}}$ and $T_{\rm{img} \gets \rm{cam}}$ are transformation matrixes from the coordinate of LiDAR to camera and the coordinate of camera to image, respectively. Instead of sampling image features $v^k_{\rm{img}}$ in the feature map with corresponding scale, we sample them in the finest feature map $I_0$ for it contains rich semantics with high resolution, which is important to detect distant objects. It is written as
\begin{equation}
    v^k_{\rm{img}} = G\left(I_0, p^k_{\rm{img}}\right),
\end{equation}
where $G(\cdot, \cdot)$ denotes the grid sampler that samples features of a position on a feature map with bilinear interpolation. Finally, we concatenate $v^k$ and $v^k_{\rm{img}}$ and employ an MLP to aggregate these features, which is formulated as
\begin{equation}
    v^k_{\rm{fuse}} = {\rm{MLP}}\left({\rm{CONCAT}}\left(v^k, v^k_{\rm{img}}\right)\right),
\end{equation}
where $v^k_{\rm{fuse}}$ denotes the fused features which share the same channels with $v^k$. Although there are other fusion manners, \emph{e.g.}, simple attention~\cite{huang2020epnet} or cross attention~\cite{zhang2022cat}, we select concatenation for efficiency. Since the regular sparse convolution would dilate the features of occupied voxels to their neighbors, $P^k$ changes along with $k$. It means that voxels in different scales may fetch different image features, and thus much more image features would be sampled in total.

\subsection{Featured-Cued Confidence Rectification}
We introduce the FCR module to further reduce the information loss of images, as depicted in Fig~\ref{feature_and_decision_fusion}~(b). Considering images provide rich textural and semantic information which makes them superior in classification, FCR aims to correct the confidence of detection candidates with the help of images. However, generating 3D and 2D detection candidates separately from two branches is computation-intensive. Inspired by~\cite{pang2022fast}, we project 3D RoIs onto the image plane to get 2D RoIs and refine them respectively to obtain 3D and 2D detection candidates. Note that the 2D Region-based Convolutional Neural Network~(R-CNN) only needs to classify these 2D RoIs, instead of both classifying and regressing them. The confidence scores of a 3D detection candidate and the corresponding 2D candidate, denoted by $s_{3D}$ and $s_{2D}$ respectively, represent the probabilities LiDAR stream and camera stream predict that there is a foreground object. What we want to do is to fuse the two scores in a learnable way.  Nevertheless, it is difficult for a network to learn a more accurate score with only two scores. Therefore, we feed the features of RoIs into the network as cues to rectify the confidence scores. In more detail, denote the features of a 3D RoI and a 2D RoI by $F_{\rm{3D}}^{\rm{roi}}$ and $F_{\rm{3D}}^{\rm{roi}}$, we combine them as
\begin{equation}
    F_{\rm{fuse}}^{\rm{roi}} = {\rm{CONCAT}}\left({\rm{AvgPool}}\left(F_{\rm{3D}}^{\rm{roi}}\right), {\rm{AvgPool}}\left(F_{\rm{2D}}^{\rm{roi}}\right)\right),
\end{equation}
where ${\rm{AvgPool}}(\cdot)$ is the average pooling operation used to reduce the feature channels. To avoid the confidence rectification being dominated by the RoI features, we concatenate $s_{3D}$ and $s_{2D}$, and map them to a higher dimension
\begin{equation}
    F_{\rm{fuse}}^{\rm{score}} = {\rm{MLP}}\left({\rm{CONCAT}\left(s_{3D}, s_{2D}\right)}\right).
\end{equation}
Finally, we employ an MLP with a sigmoid activation function $\sigma$ to predict the rectified confidence score
\begin{equation}
    s_{\rm{rect}} = \sigma \left({\rm{MLP}}\left({{\rm{CONCAT}}\left(F_{\rm{fuse}}^{\rm{roi}}, F_{\rm{fuse}}^{\rm{score}}\right)}\right)\right),
\end{equation}
Given that the spatial localization ability of point clouds is much better than images, we do not exploit images to rectify the regression.

\subsection{Occlusion-Aware GT Sampling}
Data augmentation is essential to reduce the overfitting of the network. For LiDAR-based 3D object detection, GT Sampling is a frequently used data augmentation. It randomly samples several objects from the database and pastes them into the current training scene. A collision test should be performed to remove sampled objects which are heavily occluded with other objects.

As for cross-modal 3D object detection, it is necessary but hard to main consistency between point clouds and images during data augmentation. PointAugmenting~\cite{wang2021pointaugmenting} pastes point cloud-image patch pairs into the training scene and performs the collision test on both the bird's eye view~(BEV) and the image plane. Although the consistency problem is solved, most of the sampled objects would be removed for severe occlusion on the image plane, which weakens the effect of GT Sampling. To solve this problem, we devise the OGS module. Because a sampled object may be severely occluded with many other objects, if we only remove the most severely occluded sampled object, other sampled objects occluded with this object could be retained. Therefore, we calculate how many other objects each sampled object is occluded with and remove the most severely occluded sampled object one by one. We find that this simple strategy is useful to reserve more sampled objects in training scenes, especially when we train multiple categories together. The detailed implementation is shown in Algorithm \ref{oa_gtaug}. \looseness=-1

\begin{algorithm}[t!]
\caption{Occlusion-Aware GT Sampling}
\begin{algorithmic}[1]
\REQUIRE  Ground Truth Boxes $\mathbf{G=\{g_1,g_2,...,g_N\}}$, Sampled Boxes $\mathbf{S=\{s_1,s_2,...,s_M\}}$, Occlusion Threshold on BEV Plane $\tau_1$, Occlusion Threshold on Image Plane $\tau_2$.
\STATE $\mathbf{I_1} \leftarrow {\rm{CalculateBEVIoU}}(\mathbf{S}, \mathbf{S 
\bigcup G}) $
\STATE $\mathbf{I_2} \leftarrow  {\rm{CalculateImageIoU}}( \mathbf{S} ,  \mathbf{S \bigcup G})$
\STATE $\mathbf{O} \leftarrow {\rm{CalculateOcclusionNum}}(\mathbf{I_{1}}, \mathbf{I_{2}}, \tau_1, \tau_2)$
\WHILE{O $\ne$ empty}
    \STATE $//$ find and delete the most heavily occluded sampled object
    \STATE i $\leftarrow$ Argmax O
    \STATE $\mathbf{S} \leftarrow \mathbf{S - \{s_i\}}$
    \STATE $//$ remove the items related to the $i$-th sampled object in the IoU tables
    \STATE $\mathbf{I_1} \leftarrow {\rm{Remove}}(\mathbf{I_1}, i)$
    \STATE $\mathbf{I_2} \leftarrow {\rm{Remove}}(\mathbf{I_2}, i)$
    \STATE $\mathbf{O} \leftarrow {\rm{CalculateOcclusionNum}}(\mathbf{I_{1}}, \mathbf{I_{2}}, \tau_1, \tau_2)$   
\ENDWHILE
\ENSURE $\mathbf{S}$
\end{algorithmic}
\label{oa_gtaug}
\end{algorithm}

\section{Experiments}
\subsection{Dataset and Evaluation Metrics}
We evaluate our method in the KITTI 3D object detection dataset~\cite{geiger2012we}. It contains 7,481 training frames and 7,518 testing frames. Following the data split routine, the training frames are further divided into a \emph{train} set with 3,712 frames and a \emph{val} set with 3,769 frames. According to the size, occlusion, and truncation, objects are grouped into three difficulty levels, \emph{i.e.}, easy, moderate and hard. We evaluate the results by the mAP. Following~\cite{shi2020pv}, the mAP is calculated with 11 recall positions on the \emph{val} set and with 40 recall positions on the \emph{test} set. 

\subsection{Implementation Details}
We select Voxel R-CNN~\cite{deng2021voxel} as the default baseline. MLF-DET is trained in three stages. First, we 
leverage 2D labels of KITTI dataset to train the Faster R-CNN~\cite{ren2015faster} with batch size 40, learning rate 0.02 for 12 epochs. Then we load the pre-trained weights of ResNet and FPN in Faster R-CNN and train the LiDAR stream and the MVI module with batch size 16, learning rate 0.01 for 80 epochs. OGS and other widely adopted data augmentations, \emph{i.e.}, random flipping, global scaling, and global rotation, are used in this stage. Finally, we train the FCR module with batch size 48, learning rate 0.01 for 10 epochs. The Adam optimizer with cosine annealing learning rate strategy is adopted in the second and third stages. All the experiments are done with 4 RTX 2080TI GPUs. Note that we do not use any extra data for training or testing.

\begin{table}[t!]
\centering
\caption{Performance comparison for three classes with state-of-the-art 3D detectors on the KITTI \emph{test} set. L and I in the Modality column refer to LiDAR and image, respectively. DM, RI, SS and IS in the Extra Data column refer to depth map, right image and semantic segmentation label and instance segmentation label, respectively. The best results without using extra data are highlighted in bold and the best results using extra data are underlined.}
\label{kitti_test}
\resizebox{\linewidth}{!}{
\begin{tabular}{c|c|c|c|ccc|ccc|ccc|c}
    \hline
    \multicolumn{1}{c|}{ \multirow{2}{*}{Method}}
    & \multicolumn{1}{c|}{ \multirow{2}{*}{Reference}}
    & \multicolumn{1}{c|}{ \multirow{2}{*}{Modality}}
    & \multicolumn{1}{c|}{ \multirow{2}{*}{Extra Data}}
    & \multicolumn{3}{c|}{Car~(IoU=0.7)}
    & \multicolumn{3}{c|}{Pedetrian~(IoU=0.5)}
    & \multicolumn{3}{c|}{Cyclist~(IoU=0.5)}
    & \multirow{2}{*}{mAP} \\
    \cline{5-13}
    \multicolumn{1}{c|}{}
    & \multicolumn{1}{c|}{}
    & \multicolumn{1}{c|}{}
    & \multicolumn{1}{c|}{}
    & \multicolumn{1}{c}{Easy} & \multicolumn{1}{c}{Mod.} & \multicolumn{1}{c|}{Hard} 
    & \multicolumn{1}{c}{Easy} & \multicolumn{1}{c}{Mod.} & \multicolumn{1}{c|}{Hard} 
    & \multicolumn{1}{c}{Easy} & \multicolumn{1}{c}{Mod.} & \multicolumn{1}{c|}{Hard}
    & \multicolumn{1}{c}{}\\
    \hline
    SECOND~\cite{yan2018second} & Sensors 2018 & L & - & 84.65 & 75.96 & 68.71 & 45.31 & 35.52 & 33.14 & 75.83 & 60.82 & 53.67 & 59.29 \\
    PointRCNN~\cite{shi2019pointrcnn} & CVPR 2019 & L & - & 86.96 & 75.64 & 70.70 & 47.98 & 39.37 & 36.01 & 74.96 & 58.82 & 52.53 & 60.33\\
    PointPillars~\cite{lang2019pointpillars} & CVPR 2019 & L & - & 82.58 & 74.31 & 68.99 & 51.45 & 41.92 & 38.89 & 77.10 & 58.65 & 51.92 & 60.65 \\
    Part-$A^2$~\cite{shi2020points} & TPAMI 2020 & L & - & 87.81 & 78.49 & 73.51 & 53.10 & 43.35 & 40.06 & 79.17 & 63.52 & 56.93 & 63.99 \\
    PV-RCNN~\cite{shi2020pv} & CVPR 2020 & L & - & 90.25 & 81.43 & 76.82 & 52.17 & 43.29 & 40.29 & 78.60 & 63.71 & 57.65 & 64.91 \\
    Voxel R-CNN~\cite{deng2021voxel} & AAAI 2021 & L & - & 90.90 & 81.62 & 77.06 & - & - & - & - & - & - & -\\
    IA-SSD~\cite{zhang2022not} & CVPR 2022 & L & - & 88.87 & 80.32 & 75.10 & 49.01 & 41.20 & 38.03 & 80.78 & 66.01 & 58.12 & 64.16 \\
    FocalsConv~\cite{chen2022focal} & CVPR 2022 & L & - & 90.20 & 82.12 & 77.50 & - & - & - & - & - & - & -\\
    \hline
    EPNet~\cite{huang2020epnet} & ECCV 2020 & L+I & - & 89.81 & 79.28 & 74.59 & - & - & - & - & - & - & -\\
    CLOCs-PVCas~\cite{pang2020clocs} & IROS 2020 & L+I & - & 88.94 & 80.67 & 77.15 & - & - & - & - & - & - & -\\
    Fast-CLOCs-PV~\cite{pang2022fast} & WACV 2022 & L+I & - & 89.11 & 80.34 & 76.98 & 52.10 & 42.72 & 39.08 & 82.83 & 65.31 & 57.43 & 65.10\\
    FocalsConv-F~\cite{chen2022focal} & CVPR 2022 & L+I & - & 90.55 & 82.28 & 77.59 & - & - & - & - & - & - & -\\
    VFF~\cite{li2022voxel} & CVPR 2022 & L+I & - & 89.50 & 82.09 & 79.29 & - & - & - & - & - & - & -\\
    CAT-DET~\cite{zhang2022cat} & CVPR 2022 & L+I & - & 89.87 & 81.32 & 76.68 & \textbf{54.26} & \textbf{45.44} & 41.94 & 83.68 & 68.81 & 61.45 & 67.05 \\
    HMFI~\cite{li2022homogeneous} & ECCV 2022 & L+I & - & 88.90 & 81.93 & 77.30 & 50.88 & 42.65 & 39.78 & \textbf{84.02} & 70.37 & 62.57 & 66.49\\
    DVF-PV~\cite{mahmoud2023dense} & WACV 2023 & L+I & - & 90.99 & 82.40 & 77.37 & - & - & - & - & - & - & -\\ 
    \hline
    PointPainting~\cite{vora2020pointpainting} & CVPR 2020 & L+I & SS & 82.11 & 71.70 & 67.08 & 50.32 & 40.97 & 37.84 & 77.63 & 63.78 & 55.89 & 60.81\\
    EPNet++~\cite{liu2022epnet++} & TPAMI 2022 & L+I & IS & 91.37 & 81.96 & 76.71 & 52.79 & 44.38 & 41.29 & 76.15 & 59.71 & 53.67 & 64.23\\
    VPFNet~\cite{zhu2022vpfnet} & TMM 2022 & L+I & RI+IS & 91.02 & 83.21 & 78.20 & - & - & - & - & - & - & -\\
    SFD~\cite{wu2022sparse} & CVPR 2022 & L+I & DM & \underline{91.73} & \underline{84.76} & 77.92 & - & - & - & - & - & - & -\\
    BiProDet~\cite{zhang2023bidirectional} & ICLR 2023 & L+I & IS & 89.13 & 82.97 & \underline{80.05} & \underline{55.59} & \underline{48.77} & \underline{46.12} & \underline{86.74} & \underline{74.32} & \underline{67.45} & \underline{70.13}\\
    \hline
    \textbf{MLF-DET~(ours)} & - & L+I & - & \textbf{91.18} & \textbf{82.89} & \textbf{77.89} & 50.86 & 45.29 & \textbf{42.05} & 83.31 & \textbf{70.71} & \textbf{63.71} & \textbf{67.54} \\
    \hline
\end{tabular}
}
\end{table}

\subsection{Comparison with State-of-the-Arts}
We compare our MLF-DET with other LiDAR-only and cross-modal state-of-the-art methods on both the \emph{test} and \emph{val} sets. We show the input modality and the extra data used in the tables for each method for a fair comparison. It is worth noting that many previous approaches train different models for different categories to get better performance, which we argue is meaningless for practical application. Instead, we train our network for the three classes and report the results on a single model. Besides, we do not employ any test time augmentation or model ensemble, so as to provide the actual performance of our method.

\noindent\textbf{Comparison on KITTI \emph{test} set.}
As represented in Table~\ref{kitti_test}, MLF-DET achieves competitive or  superior performance over recently published state-of-the-art methods and obtains the highest mAP among the methods that do not use extra data. Compared with our LiDAR-only baseline Voxel R-CNN, we achieve up to 1.27\% and 0.83\% gains in AP$_{Mod.}$ and AP$_{Hard}$ of the car class, respectively, indicating the effectiveness of our fusion scheme. Although some methods perform better than ours, they need extra  sensor data or annotations, which limits their real-world deployment. By contrast, our approach only needs the LiDAR point clouds and monocular images as input and adopts the 3D and 2D bounding box annotations as training labels.

\begin{figure}[t!]
\centering
\includegraphics[width=\textwidth]{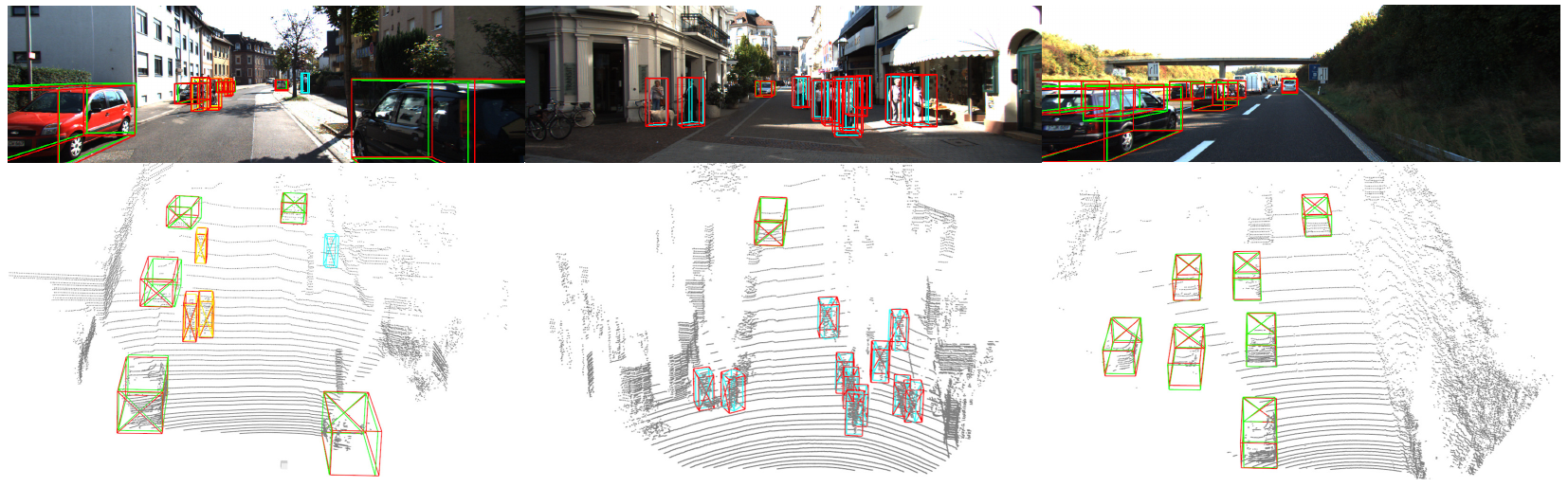}
\caption{Qualitative results on KITTI \emph{val} set. Green, cyan and yellow boxes indicate predicted objects of car, pedestrian and cyclist classes, respectively. Red boxes indicate ground truth objects. Best viewed in color.} 
\label{visualization}
\end{figure}

\noindent\textbf{Comparison on KITTI \emph{val} set.}
As shown in Table~\ref{kitti_val}, MLF-DET also achieves the best mAP compared with the methods not using extra data in the \emph{val} set. Remarkably, MLF-DET-V gets 87.31\% and 68.50\% AP$_{Mod.}$ for car and pedestrian, which even outperform methods using extra data. To show the generalization capability of our fusion scheme, we also select PV-RCNN~\cite{shi2020pv} as the baseline. MLF-DET-PV surpasses PV-RCNN consistently in different categories and different difficulty levels and gives a 2.98\% mAP gain. In addition, we provide the qualitative results of our method in Fig~\ref{visualization}. It is shown that MLF-DET can accurately detect objects of different classes, despite some of them being severely occluded or far away from the ego-vehicle. In the first column  of Fig~\ref{visualization}, we surprisingly find that MLF-DET can even detect a pedestrian with few LiDAR points that is ignored by human annotators.  

\begin{table}[t!]
\centering
\caption{Performance comparison with state-of-the-art 3D detectors on the KITTI \emph{val} set. MLF-DET-PV and MLF-DET-V denote MLF-DET based on PV-RCNN and Voxel R-CNN, respectively.}
\label{kitti_val}
\resizebox{\linewidth}{!}{
\begin{tabular}{c|c|c|c|ccc|ccc|ccc|c}
    \hline
    \multicolumn{1}{c|}{ \multirow{2}{*}{Method}}
    & \multicolumn{1}{c|}{ \multirow{2}{*}{Reference}}
    & \multicolumn{1}{c|}{ \multirow{2}{*}{Modality}}
    & \multicolumn{1}{c|}{ \multirow{2}{*}{Extra Data}}
    & \multicolumn{3}{c|}{Car~(IoU=0.7)}
    & \multicolumn{3}{c|}{Pedetrian~(IoU=0.5)}
    & \multicolumn{3}{c|}{Cyclist~(IoU=0.5)}
    & \multirow{2}{*}{mAP} \\
    \cline{5-13}
    \multicolumn{1}{c|}{}
    & \multicolumn{1}{c|}{}
    & \multicolumn{1}{c|}{}
    & \multicolumn{1}{c|}{}
    & \multicolumn{1}{c}{Easy} & \multicolumn{1}{c}{Mod.} & \multicolumn{1}{c|}{Hard} 
    & \multicolumn{1}{c}{Easy} & \multicolumn{1}{c}{Mod.} & \multicolumn{1}{c|}{Hard} 
    & \multicolumn{1}{c}{Easy} & \multicolumn{1}{c}{Mod.} & \multicolumn{1}{c|}{Hard}
    & \multicolumn{1}{c}{}\\
    \hline
    SECOND~\cite{yan2018second} & Sensors 2018 & L & - & 88.61 & 78.62 & 77.22 & 56.55 & 52.98 & 47.73 & 80.58 & 67.15 & 63.10 & 68.06 \\
    PointRCNN~\cite{shi2019pointrcnn} & CVPR 2019 & L & - & 88.72 & 78.61 & 77.82 & 62.72 & 53.85 & 50.24 & 86.84 & 71.62 & 65.59 & 70.67\\
    Part-$A^2$~\cite{shi2020points} & TPAMI 2020 & L & - & 89.55 & 79.40 & 78.84 & 65.68 & 60.05 & 55.44 & 85.50 & 69.90 & 65.48 & 72.20\\
    PV-RCNN~\cite{shi2020pv} & CVPR 2020 & L & - & 89.03 & 83.24 & 78.59 & 63.71 & 57.37 & 52.84 & 86.06 & 69.48 & 64.50 & 71.65\\
    Voxel R-CNN~\cite{deng2021voxel} & AAAI 2021 & L & - & 89.41 & 84.52 & 78.93 & - & - & - & - & - & - & - \\
    FocalsConv~\cite{chen2022focal} & CVPR 2022 & L & - & 89.52 & 84.93 & 79.18 & - & - & - & - & - & - & -\\
    \hline
    EPNet~\cite{huang2020epnet} & ECCV 2020 & L+I & - & 88.76 & 78.65 & 78.32 & 66.74 & 59.29 & 54.82 & 83.88 & 65.50 & 62.70 & 70.96 \\
    CLOCs-PVCas~\cite{pang2020clocs} & IROS 2020 & L+I & - & 89.49 & 79.31 & 77.36 & 62.88 & 56.20 & 50.10 & 87.57 & 67.92 & 63.67 & 70.50\\
    FocalsConv-F~\cite{chen2022focal} & CVPR 2022 & L+I &  & 89.82 & 85.22 & 85.19 & - & - & - & - & - & - & - \\
    VFF~\cite{li2022voxel} & CVPR 2022 & L+I & - & 89.51 & 84.76 & 79.21 & -  & - & - & - & - & - & -\\
    CAT-DET~\cite{zhang2022cat} & CVPR 2022 & L+I & - & \textbf{90.12} & 81.46 & 79.15 & \textbf{74.08} & 66.35 & 58.92 & \textbf{87.64} & \textbf{72.82} & \textbf{68.20} & 75.42 \\
    HMFI~\cite{li2022homogeneous} & ECCV 2022 & L+I & - & - & 85.14 & - & - & 62.41 & - & - & 74.11 & - & -\\
    \hline
    SFD~\cite{wu2022sparse} & CVPR 2022 & L+I & DM & \underline{89.74} & \underline{87.12} & \underline{85.20} & -  & - & - & - & - & - & -\\
    BiProDet~\cite{zhang2023bidirectional} & ICLR 2023 & L+I & IS & 89.73 & 86.40 & 79.31 & \underline{71.77} & \underline{68.49} & \underline{62.52} & \underline{89.24} & \underline{76.91} & \underline{75.18} & \underline{77.73}\\
    \hline
    \textbf{MLF-DET-PV~(ours)} & - & L+I & - & 89.48 & 86.71 & 79.33 & 68.99 & 61.94 & 59.55 & 87.16 & 72.53 & 66.02 & 74.63 \\
    \textbf{MLF-DET-V~(ours)} & - & L+I & - & 89.70 & \textbf{87.31} & \textbf{79.34} & 71.15 & \textbf{68.50} & \textbf{61.72} & 86.05 & 72.14 & 65.42 & \textbf{75.70} \\
    \hline
\end{tabular}
}
\end{table}

\subsection{Ablation Study}
To analyze the effectiveness of the individual components of our method, we conduct an ablation study on the KITTI \emph{val} set. Since the proportion of the pedestrian and cyclist classes is much smaller than that of the car class and their performance is unstable, we only report the recall of RoIs and AP on the car class. As shown in Table~\ref{ablation_stady}, our MLF-DET can bring a 2.57\% gain in recall with IoU 0.7 and a 3.09\% gain in AP$_{Mod.}$.

\begin{table}[t!]
\scriptsize
\centering
\caption{Effects of three proposed modules for car class on the KITTI \emph{val} set. $\tau$ denotes the 3D IoU threshold of ground truth boxes and predicted boxes. $^\dag$ means the re-implemented results with the official code on our device. }
\label{ablation_stady}
\begin{tabular}{c|ccc|cc|cccc}
    \hline
    \multicolumn{1}{c|}{ \multirow{2}{*}{Method}}  & \multicolumn{1}{c}{ \multirow{2}{*}{MVI}}
    & \multicolumn{1}{c}{ \multirow{2}{*}{OGS}} & \multicolumn{1}{c|}{ \multirow{2}{*}{FCR}} & \multicolumn{2}{c|}{Recall} & \multicolumn{4}{c}{Average Precision}\\
    \cline{5-10}
    \multicolumn{1}{c|}{} 
    & \multicolumn{1}{c}{}
    & \multicolumn{1}{c}{}
    & \multicolumn{1}{c|}{}
    & \multicolumn{1}{c}{$\tau$=0.5} & \multicolumn{1}{c|}{$\tau$=0.7}
    & \multicolumn{1}{c}{Easy} & \multicolumn{1}{c}{Mod.} & \multicolumn{1}{c}{Hard} & \multicolumn{1}{c}{mAP}\\
    \hline
     Baseline$^\dag$~\cite{deng2021voxel} & - & - & - & 97.14 & 86.01 & 89.47 & 84.22 & 78.77 & 84.15 \\
    \hline
    \multirow{4}{*}{Ours} & - & - & $\surd$ & 97.14 & 86.01 & 89.49 & 86.25 & 78.98 & 84.91 \\
    & $\surd$ & - & - & 97.18 & 87.66 & 89.65 & 86.42 & 79.13 & 85.07 \\
    & $\surd$ & $\surd$ & - & 97.44 & 88.58 & 89.69 & 86.57 & 79.23 & 85.16 \\
    & $\surd$ & $\surd$ & $\surd$ & 97.44 & 88.58 & 89.70 & 87.31 & 79.34 & 85.45 \\
    \hline
    \multicolumn{4}{c|}{Improvements} & +0.30 & +2.57 & +0.23 & +3.09 & +0.57 & +1.30\\
    \hline
\end{tabular}
\end{table}

\begin{table}[t]
\centering
\caption{Inference speed comparison with other cross-modal 3D detectors.}
\label{speed}
\resizebox{\linewidth}{!}{
\begin{tabular}{c|c|c|c|c|c|c}
\hline
PointPainting~\cite{vora2020pointpainting} & VPFNet~\cite{zhu2022vpfnet} & SFD~\cite{wu2022sparse} & CAT-DET~\cite{zhang2022cat} & BiProDet~\cite{zhang2023bidirectional} & HMFI~\cite{li2022homogeneous} & MLF-DET \\
\hline
2.5 FPS & 5.9 FPS & 15.0 FPS & 10.2 FPS & 3.3 FPS & 9.5 FPS & 10.8 FPS \\
\hline
\end{tabular}
}
\end{table}

\noindent\textbf{Effect of Multi-Scale Voxel Image Fusion.}
When MVI module is applied to the baseline, the recall with IoU 0.7 is improved by 1.65\% and AP$_{Mod.}$ is enhanced by 2.20\%. We attribute the improvements to the dense feature-level fusion scheme that fetches rich semantic features from the high-resolution image feature map. With these features, the network is able to detect some distant objects with few LiDAR points, but it is hard for the LiDAR-only baseline.

\noindent\textbf{Effect of Feature-Cued Confidence Rectification.}
FCR module leads to 2.03\% and 0.74\% performance boost for the baseline and the baseline with MVI and OGS modules in AP$_{Mod.}$. The latter boost demonstrates the significance of combining feature- and decision-level fusion. The feature-level fusion enables our baseline detector to generate high-quality RoIs, but some of these RoIs could not be correctly classified. By rectifying the scores of RoIs, the FCR module reduces false positives and false negatives, thus enhancing the AP. Since the FCR module only operates on the scores of RoIs, there are no gains in the recall.

\begin{wrapfigure}{r}{4.5cm}
	\centering
	\vskip -0.34 in
	\includegraphics[width=4.5cm]{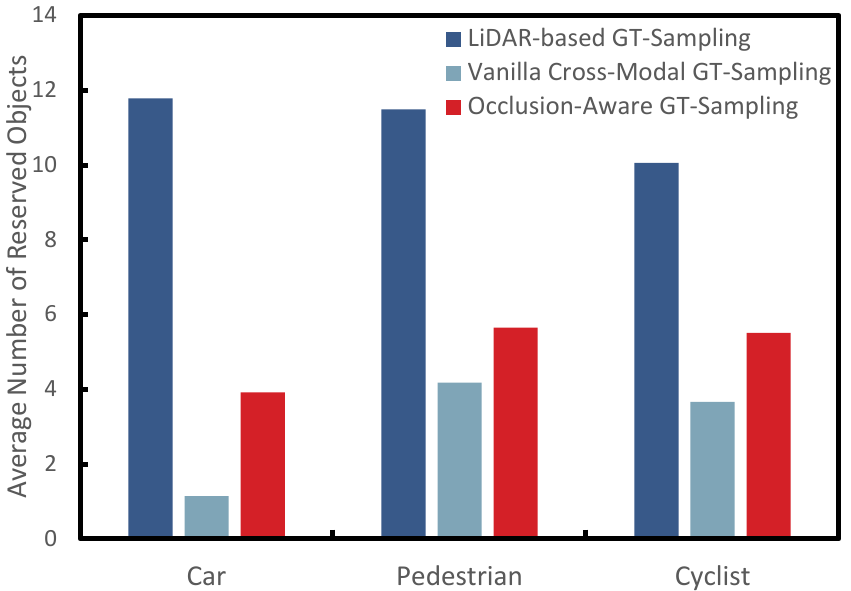}
	\vskip -0.20 in
	\caption{Comparison in the average number of reserved objects among different GT Sampling methods.}
	\label{sample-num}
	\vskip -0.4 in
\end{wrapfigure}
\noindent\textbf{Effect of Occlusion-Aware GT Sampling.}
Although OGS module does not bring many improvements in the AP, it effectively increases the recall by 0.92\%. The reason is that OGS provides more sampled objects in training, forcing the network to find more objects. To further understand the effect of OGS, we calculate the average number of reserved objects of three GT Sampling methods on the \emph{train} set with a fixed random seed, as shown in Fig~\ref{sample-num}. This shows that OGS could preserve more sampled objects than vanilla cross-modal GT Sampling, especially in the car class. 

\noindent\textbf{Inference Speed.}
Since MLF-DET integrates feature- and decision-level fusion into a single framework, an intuitive concern is that the inference speed of MLF-DET would be slow. To investigate it, we run our model on an NVIDIA RTX 2080TI GPU and measure the FPS with the full GPU memory utilization. As shown in Table~\ref{speed}, MLF-DET is not very slow compared with other cross-modal methods. The reason lies in two aspects. First, the LiDAR branch of MLF-DET is lightweight and fast. Second, MLF-DET does not apply the attention mechanism to fuse cross-modal features and the RoIs are shared between LiDAR and image branches. Therefore, the RPN in the 2D detector can be removed.

\section{Conclusion}
In this paper, we propose a multi-level fusion network termed MLF-DET for cross-modal 3D detection. To fully utilize images, we design the MVI module to densely combine the features of images and point clouds, and the FCR module to rectify the scores of RoIs by fusing the scores and features of RoIs from LiDAR branch and camera branch. Additionally, we propose the OGS module to preserve more sampled objects in training. Experiments on the KITTI dataset demonstrate the accuracy of our detector and the effectiveness of these modules. In future work, we would like to explore a more reasonable architecture or training strategy to optimize the feature-level and decision-level fusion modules in an end-to-end manner.

\subsubsection{Acknowledgements} This work was supported by the National Natural Science Foundation of China under Grant No.62088102.

\bibliographystyle{splncs04}
\bibliography{reference}

\begin{thebibliography}{10}
\providecommand{\url}[1]{\texttt{#1}}
\providecommand{\urlprefix}{URL }
\providecommand{\doi}[1]{https://doi.org/#1}

\bibitem{bai2022transfusion}
Bai, X., Hu, Z., Zhu, X., Huang, Q., Chen, Y., Fu, H., Tai, C.L.: Transfusion:
  Robust lidar-camera fusion for 3d object detection with transformers. In:
  Proceedings of the IEEE/CVF conference on computer vision and pattern
  recognition. pp. 1090--1099 (2022)

\bibitem{chen2022focal}
Chen, Y., Li, Y., Zhang, X., Sun, J., Jia, J.: Focal sparse convolutional
  networks for 3d object detection. In: Proceedings of the IEEE/CVF Conference
  on Computer Vision and Pattern Recognition. pp. 5428--5437 (2022)

\bibitem{deng2021voxel}
Deng, J., Shi, S., Li, P., Zhou, W., Zhang, Y., Li, H.: Voxel r-cnn: Towards
  high performance voxel-based 3d object detection. In: Proceedings of the AAAI
  Conference on Artificial Intelligence. vol.~35, pp. 1201--1209 (2021)

\bibitem{geiger2012we}
Geiger, A., Lenz, P., Urtasun, R.: Are we ready for autonomous driving? the
  kitti vision benchmark suite. In: 2012 IEEE conference on computer vision and
  pattern recognition. pp. 3354--3361. IEEE (2012)

\bibitem{graham20183d}
Graham, B., Engelcke, M., Van Der~Maaten, L.: 3d semantic segmentation with
  submanifold sparse convolutional networks. In: Proceedings of the IEEE
  conference on computer vision and pattern recognition. pp. 9224--9232 (2018)

\bibitem{he2016deep}
He, K., Zhang, X., Ren, S., Sun, J.: Deep residual learning for image
  recognition. In: Proceedings of the IEEE conference on computer vision and
  pattern recognition. pp. 770--778 (2016)

\bibitem{huang2020epnet}
Huang, T., Liu, Z., Chen, X., Bai, X.: Epnet: Enhancing point features with
  image semantics for 3d object detection. In: Computer Vision--ECCV 2020: 16th
  European Conference, Glasgow, UK, August 23--28, 2020, Proceedings, Part XV
  16. pp. 35--52. Springer (2020)

\bibitem{lang2019pointpillars}
Lang, A.H., Vora, S., Caesar, H., Zhou, L., Yang, J., Beijbom, O.:
  Pointpillars: Fast encoders for object detection from point clouds. In:
  Proceedings of the IEEE/CVF conference on computer vision and pattern
  recognition. pp. 12697--12705 (2019)

\bibitem{li2020rtm3d}
Li, P., Zhao, H., Liu, P., Cao, F.: Rtm3d: Real-time monocular 3d detection
  from object keypoints for autonomous driving. In: Computer Vision--ECCV 2020:
  16th European Conference, Glasgow, UK, August 23--28, 2020, Proceedings, Part
  III 16. pp. 644--660. Springer (2020)

\bibitem{li2022homogeneous}
Li, X., Shi, B., Hou, Y., Wu, X., Ma, T., Li, Y., He, L.: Homogeneous
  multi-modal feature fusion and interaction for 3d object detection. In:
  Computer Vision--ECCV 2022: 17th European Conference, Tel Aviv, Israel,
  October 23--27, 2022, Proceedings, Part XXXVIII. pp. 691--707. Springer
  (2022)

\bibitem{li2022voxel}
Li, Y., Qi, X., Chen, Y., Wang, L., Li, Z., Sun, J., Jia, J.: Voxel field
  fusion for 3d object detection. In: Proceedings of the IEEE/CVF Conference on
  Computer Vision and Pattern Recognition. pp. 1120--1129 (2022)

\bibitem{lin2017feature}
Lin, T.Y., Doll{\'a}r, P., Girshick, R., He, K., Hariharan, B., Belongie, S.:
  Feature pyramid networks for object detection. In: Proceedings of the IEEE
  conference on computer vision and pattern recognition. pp. 2117--2125 (2017)

\bibitem{liu2020smoke}
Liu, Z., Wu, Z., T{\'o}th, R.: Smoke: Single-stage monocular 3d object
  detection via keypoint estimation. In: Proceedings of the IEEE/CVF Conference
  on Computer Vision and Pattern Recognition Workshops. pp. 996--997 (2020)

\bibitem{liu2022epnet++}
Liu, Z., Huang, T., Li, B., Chen, X., Wang, X., Bai, X.: Epnet++: Cascade
  bi-directional fusion for multi-modal 3d object detection. IEEE Transactions
  on Pattern Analysis and Machine Intelligence  (2022)

\bibitem{mahmoud2023dense}
Mahmoud, A., Hu, J.S., Waslander, S.L.: Dense voxel fusion for 3d object
  detection. In: Proceedings of the IEEE/CVF Winter Conference on Applications
  of Computer Vision. pp. 663--672 (2023)

\bibitem{pang2020clocs}
Pang, S., Morris, D., Radha, H.: Clocs: Camera-lidar object candidates fusion
  for 3d object detection. In: 2020 IEEE/RSJ International Conference on
  Intelligent Robots and Systems (IROS). pp. 10386--10393. IEEE (2020)

\bibitem{pang2022fast}
Pang, S., Morris, D., Radha, H.: Fast-clocs: Fast camera-lidar object
  candidates fusion for 3d object detection. In: Proceedings of the IEEE/CVF
  Winter Conference on Applications of Computer Vision. pp. 187--196 (2022)

\bibitem{qi2019deep}
Qi, C.R., Litany, O., He, K., Guibas, L.J.: Deep hough voting for 3d object
  detection in point clouds. In: proceedings of the IEEE/CVF International
  Conference on Computer Vision. pp. 9277--9286 (2019)

\bibitem{qi2017pointnet}
Qi, C.R., Su, H., Mo, K., Guibas, L.J.: Pointnet: Deep learning on point sets
  for 3d classification and segmentation. In: Proceedings of the IEEE
  conference on computer vision and pattern recognition. pp. 652--660 (2017)

\bibitem{qi2017pointnet++}
Qi, C.R., Yi, L., Su, H., Guibas, L.J.: Pointnet++: Deep hierarchical feature
  learning on point sets in a metric space. Advances in neural information
  processing systems  \textbf{30} (2017)

\bibitem{ren2015faster}
Ren, S., He, K., Girshick, R., Sun, J.: Faster r-cnn: Towards real-time object
  detection with region proposal networks. Advances in neural information
  processing systems  \textbf{28} (2015)

\bibitem{shi2020pv}
Shi, S., Guo, C., Jiang, L., Wang, Z., Shi, J., Wang, X., Li, H.: Pv-rcnn:
  Point-voxel feature set abstraction for 3d object detection. In: Proceedings
  of the IEEE/CVF Conference on Computer Vision and Pattern Recognition. pp.
  10529--10538 (2020)

\bibitem{shi2019pointrcnn}
Shi, S., Wang, X., Li, H.: Pointrcnn: 3d object proposal generation and
  detection from point cloud. In: Proceedings of the IEEE/CVF conference on
  computer vision and pattern recognition. pp. 770--779 (2019)

\bibitem{shi2020points}
Shi, S., Wang, Z., Shi, J., Wang, X., Li, H.: From points to parts: 3d object
  detection from point cloud with part-aware and part-aggregation network. IEEE
  transactions on pattern analysis and machine intelligence  \textbf{43}(8),
  2647--2664 (2020)

\bibitem{vora2020pointpainting}
Vora, S., Lang, A.H., Helou, B., Beijbom, O.: Pointpainting: Sequential fusion
  for 3d object detection. In: Proceedings of the IEEE/CVF conference on
  computer vision and pattern recognition. pp. 4604--4612 (2020)

\bibitem{wang2021pointaugmenting}
Wang, C., Ma, C., Zhu, M., Yang, X.: Pointaugmenting: Cross-modal augmentation
  for 3d object detection. In: Proceedings of the IEEE/CVF Conference on
  Computer Vision and Pattern Recognition. pp. 11794--11803 (2021)

\bibitem{wu2022sparse}
Wu, X., Peng, L., Yang, H., Xie, L., Huang, C., Deng, C., Liu, H., Cai, D.:
  Sparse fuse dense: Towards high quality 3d detection with depth completion.
  In: Proceedings of the IEEE/CVF Conference on Computer Vision and Pattern
  Recognition. pp. 5418--5427 (2022)

\bibitem{yan2018second}
Yan, Y., Mao, Y., Li, B.: Second: Sparsely embedded convolutional detection.
  Sensors  \textbf{18}(10), ~3337 (2018)

\bibitem{zhang2022cat}
Zhang, Y., Chen, J., Huang, D.: Cat-det: Contrastively augmented transformer
  for multi-modal 3d object detection. In: Proceedings of the IEEE/CVF
  Conference on Computer Vision and Pattern Recognition. pp. 908--917 (2022)

\bibitem{zhang2022not}
Zhang, Y., Hu, Q., Xu, G., Ma, Y., Wan, J., Guo, Y.: Not all points are equal:
  Learning highly efficient point-based detectors for 3d lidar point clouds.
  In: Proceedings of the IEEE/CVF Conference on Computer Vision and Pattern
  Recognition. pp. 18953--18962 (2022)

\bibitem{zhang2023bidirectional}
Zhang, Y., Zhang, Q., Hou, J., Yuan, Y., Xing, G.: Bidirectional propagation
  for cross-modal 3d object detection. arXiv preprint arXiv:2301.09077  (2023)

\bibitem{zhang2021objects}
Zhang, Y., Lu, J., Zhou, J.: Objects are different: Flexible monocular 3d
  object detection. In: Proceedings of the IEEE/CVF Conference on Computer
  Vision and Pattern Recognition. pp. 3289--3298 (2021)

\bibitem{zhu2022vpfnet}
Zhu, H., Deng, J., Zhang, Y., Ji, J., Mao, Q., Li, H., Zhang, Y.: Vpfnet:
  Improving 3d object detection with virtual point based lidar and stereo data
  fusion. IEEE Transactions on Multimedia  (2022)

\end{thebibliography}
\end{document}